\documentclass[lettersize,journal]{IEEEtran}
\usepackage{amsmath,amsfonts}
\usepackage{algorithmic}
\usepackage{algorithm}
\usepackage{array}
\usepackage{graphicx}
\usepackage{amsmath}
\usepackage{amssymb}
\usepackage{booktabs}
\usepackage{adjustbox}
\usepackage[caption=false,font=normalsize,labelfont=sf,textfont=sf]{subfig}
\usepackage{textcomp}
\usepackage{stfloats}
\usepackage{url}
\usepackage{verbatim}
\usepackage{graphicx}
\usepackage{cite}
\usepackage{graphicx}
\usepackage{amsmath}
\usepackage{amssymb}
\usepackage{booktabs}
\usepackage{adjustbox}
\usepackage{comment}
\usepackage[accsupp]{axessibility}  
\usepackage{fancyhdr}

\begin{document}

\title{Bimodal SegNet: Fused Instance Segmentation using Events and RGB Frames for Robotic Grasping}

\author{Sanket Kachole$^{1}$, Xiaoqian Huang$^{2}$, Fariborz Baghaei Naeini$^{1}$, Rajkumar Muthusamy$^{3}$, \\ Dimitrios Makris$^{1}$ and Yahya Zweiri$^{2}$
\thanks{*This work was supported by Kingston University, the Advanced Research and Innovation Center (ARIC), and Khalifa University of Science and Technology, Abu Dhabi, UAE.}
\thanks{$^{1}$Sanket Kachole, Fariborz Baghaei Naeini, and Dimitrios Makris are with Faculty of Science, Engineering and Computing, Kingston University, London, KT1 2EE.
        {\tt\small Corresponding author: K1742163@Kingston.ac.uk}}%
\thanks{$^{2}$Xiaoqian Huang and Yahya Zweiri are with the Aerospace Engineering Department, Khalifa University of Science and Technology, Abu Dhabi, UAE.
        {\tt\small yahya.zweiri@ku.ac.ae}}%
\thanks{$^{3}$Rajkumar Muthusamy is with Dubai Future Labs - RLab, Dubai, UAE
        {\tt\small rajkumar.muthusamy@dubaifuture.gov.ae}}%
}



\maketitle

\begin{abstract}


Object segmentation enhances robotic grasping by aiding object identification. Complex environments and dynamic conditions pose challenges such as occlusion, low light conditions, motion blur and object size variance. To address these challenges, we propose a Bimodal SegNet that fuses two types of visual signals, event-based data and RGB  frame data. The proposed Bimodal SegNet network has two distinct encoders - one for RGB signal input and another for Event signal input, in addition to an Atrous Pyramidal Feature Amplification module. Encoders capture and fuse the rich contextual information from different resolutions via a Cross-Domain Contexual Attention layer while the decoder obtains sharp object boundaries. The evaluation of the proposed method undertakes five unique image degradation challenges including occlusion, blur, brightness, trajectory and scale variance on the Event-based Segmentation (ESD) Dataset. The results show a 4-6\% segmentation accuracy improvement over state-of-the-art methods in terms of mean intersection over the union and pixel accuracy.

\end{abstract}


\section{Introduction}

Robotic grasping is the ability of robots to hold and manipulate target objects using their arms and end effectors. The grasping mechanism is mainly driven using sensors to locate the position and properties of the object for deploying advanced algorithms to control the movement of the robot's joints and fingers for a firm grasp \cite{Dinakaran2023AGrasping}. It has wider industrial applications such as manufacturing plants, assembly lines, machining tasks and warehouses. An essential requirement to attempt a successful grasp include sensing, planning, control, adaptability and robustness. Vision-based robotic grasping methods including object detection, tracking, monitoring, inspection and gripping have delivered promising results. These methods require a highly accurate perception of the environment for the robot to execute the successful actions. Thus, the quality of the collected visual information is key to improving the performance in grasping applications \cite{Du2021Vision-basedReview}. 



Although extensive efforts have been taken to deploy human-level grasping skills into robots \cite{sanket20163Tracker}, robotic grasps are still inferior to human ones. While grasping systems may operate under strict conditions (e.g. fix lighting, object type/size/orientation, etc), generalizing their operation to a wider set of environmental settings is benefited by using visual sensors which provide richer and more detailed information regarding the object to be grasped, especially when combined with machine learning methods. In particular, the application of object segmentation methods holds considerable potential for enhancing the generalizability of robotic grasping across various industrial settings. Yet, despite its promise, the utility of these methods is often limited by the inherent drawbacks of conventional RGB cameras. Specifically, visual information collected by these cameras in industrial environments may be compromised by factors such as low light conditions, motion blur, object size variance, and occlusion \cite{Bader2018ChallengesManufacturing}.


In response to these challenges, the advent of neuromorphic cameras, a recent advance in visual sensors, offers a promising avenue. Standard RGB cameras capture a full projection of the scene in sequences of frames. On the other side, neuromorphic cameras record changes in pixel brightness as a stream of “events" \cite{Liao2021NeuromorphicPerspectives}. Their advantages over standard RGB cameras is their high temporal resolution, low motion blur in high speed, low transmission bandwidth, high dynamic range and low power consumption. However, since events only record changes, their representative power is higher in textured regions  and boundaries, but lower in uniform regions \cite{Huang2020NeuromorphicApplications}, \cite{Muthusamy2021NeuromorphicServoing}. In addition, RGB cameras can capture contextual information and perform better when stationary and worse in high-speed due to motion blurring, while neuromorphic cameras are benefited by the camera motion that causes events to be generated around the object boundaries \cite{Muthusamy2020NeuromorphicManipulation}. Both sensors are thus complementary and fusing their signals within a deep learning model may improve the accuracy of instance segmentation and address multiple challenges such as occlusion, blur, brightness and scale variance \cite{Gehrig2021CombiningPrediction, Qingyun2022Cross-modalityImagery}.

To achieve this goal, we introduce the Bimodal SegNet, a deep learning network specifically designed for segmentation tasks using events and RGB frames. This model leverages the conventional U-Net architecture \cite{Ronneberger2015U-Net:Segmentation} and integrates the cross-attention mechanism from the transformer architecture \cite{Liu2022CMX:Transformers} to accommodate signals from multiple modalities and improve the segmentation results in challenging industrial settings. The proposed network features the Integrated Multisensory Encoder, which consists of two distinct encoders that correspond to each signal type. These encoders utilize a Cross Domain Contextual Attention mechanism for feature fusion. In addition, the network incorporates a module called Atrous Pyramidal Feature Amplification. This module, which is based on spatial pyramidal pooling with atrous convolutions\cite{Chen2018Encoder-DecoderSegmentation}, captures rich contextual information by pooling the weighted concatenated features at different resolutions. This facilitates the decoder in obtaining sharp object boundaries.

The main contributions of this paper are as follows:
\begin{enumerate}



\item This paper presents Bimodal SegNet, an innovative encoder-decoder-based architecture, integrating cross-attention mechanisms from transformer architecture into a conventional U-Net structure for multi-modal signal processing.

\item Our research introduces a dual-encoder approach for bimodal input, utilizing weighted features and Atrous spatial pyramidal pooling in a Bimodal encoder-decoder structure. This innovative design enables effective and robust processing of visual information across diverse conditions.

\item  Rigorous evaluation in real-world industrial applications, specifically on the ESD dataset, reveals that our proposed Bimodal SegNet model excels in performance compared to other methods, exhibiting superior results in mean intersection over union (mIoU) and pixel accuracy metrics

 \end{enumerate}


The rest of this paper is organized as follows. Section \ref{section : Related Work} reviews related works. The proposed architecture is described in detail in methodology Section \ref{section : Methodology}. Section \ref{section: Experimental Results} provides experimental results and an ablation study. Finally, Section \ref{section: Conclusion} presents the conclusion and scope for further research.


\section{{Related Work}}
\label{section : Related Work}

\subsection{\textbf{Instance Segmentation in Robotics}}
Instance segmentation, a nuanced task in computer vision, has seen significant research progress where objects in an image are not only detected but also individually segmented \cite{Kachole2020AHoneybees}. Foundational work like Mask R-CNN \cite{He2017MaskR-CNN} established a robust precedent by supplementing object detection with pixel-level identification, thus distinguishing individual object instances. Later research, namely SOLOv2\cite{Wang2020SOLOv2:Segmentation} and YOLACT\cite{Bolya2019YOLACTSegmentation}, stressed the imperative for swiftness and efficacy, designing single-stage, end-to-end processes to optimize instance segmentation implementation.

This technology's applicability in robotics, marked by its environmental perception capacity, is critical for the secure and efficient interaction of robots. Research, including \cite{Feng2022TowardsPicking} bin-picking tasks, illustrates its value in interpreting cluttered environments. Additionally, \cite{Tremblay2018DeepObjects} demonstrated the potential of combining instance segmentation with pose estimation for semantic robotic grasping in unpredictable scenarios. Instance segmentation's expansion to three-dimensional data, evidenced by \cite{Jiang2020PointGroup:Segmentation}'s PointGroup, addresses 3D point clouds, a common data type in robotics. Despite these advancements, the creation of a universally adaptable instance segmentation system suitable for various real-world robotic applications poses considerable challenges, requiring continued exploration and research \cite{Krawczyk2023SegmentationReview, Karthik2022Contour-enhancedSegmentation}.

\subsection{\textbf{Non-Traditional Modalities}}
Vision-based robotic grasping systems can be divided into mainly three types, depending on the sensors used: RGB-based,  event-based and RGB-event fusion based \cite{Ayyad2022NeuromorphicSystems,Xu2023Event-drivenModeling}. Traditional cameras capture visual information in the form of RGB frames continuously which leads to high power consumption. In addition, image quality may suffer e.g. blurring due to motion or low contrast in low lighting conditions which impact the segmentation and consequently the grasping quality \cite{Brooks2019LearningBlur}\cite{LiyuanPan2019BringingCamera}. A recent study showed that traditional cameras are incompetent to capture blur-free movement on industrial conveyor belts due to a relatively low sampling rate, compared to event-based vision systems. Besides, the robotic gripper’s actuating speed in industrial applications is a minimum of 100 ms \cite{Muthusamy2020NeuromorphicManipulation} which also contributes to motion blur. Although high-speed frame-based cameras with a frame rate of more than 100 FPS are available, they demand higher power and storage consumption. Reducing latency is a significant consideration in robotic grasping applications. Complex algorithms in computer vision and image processing consume additional computing time which eventually impacts on overall grasping process \cite{Huang2020NeuromorphicApplications, naeini2022event}. This demands faster data acquirement and processing for efficient grasping. Lastly, improvement in data capturing processing and computation can reserve additional time for the gripper’s actuation. Exploiting the advantage of high temporal resolution of events, \cite{Gehrig2018AsynchronousFrames} used the Xception encoder, supplied with 6 channels of image representation. The EV-SegNet \cite{Alonso2018EV-SegNet:Cameras} also used an Xception-based CNN to handle event data for semantic segmentation applications.  The efforts with the spiking neural network to process the event based vision modalities have been explored in the \cite{kachole2024asynchronous}.

\subsection{\textbf{Multimodal Fusion}}
Various algorithms fuse modalities to take advantage of their complementarity. DAVIS produces events and frames simultaneously, making it a suitable visual sensor. The recurrent architecture \cite{Gehrig2021CombiningPrediction} has been explored to fuse the events and frames for depth prediction. The fusion of events and RGB frames has been successfully applied in simultaneous localization and mapping \cite{Zhu2018UnsupervisedEgomotion}, high-dynamic range intensity reconstruction \cite{Tomy2017FusingConditions}, feature tracking \cite{Gehrig2018AsynchronousFrames}, and image deblurring \cite{LiyuanPan2019BringingCamera}. The integration of multimodal data has emerged as a crucial paradigm in computer vision research, offering an enriched representation of the environment compared to unimodal data. Multimodal networks leverage complementary information from various data sources such as depth, LiDAR, polarization, thermal, and event cameras, enhancing the performance of several tasks including image recognition, object detection, and segmentation \cite{Huang2023AnRecognition, Qiu2023HierarchicalFeatures}. ACNet uses the Asymmetric Convolution Block (ACB) as an architecture-neutral CNN structure to enhance rotational robustness and square convolution kernels' central parts. Its effectiveness, however, might vary across CNN architectures or datasets due to potential inconsistencies in ACB's performance improvements. Meanwhile, the SA-GATE method employs a Cross-modality Guided Encoder for recalibrating RGB features and distilling depth information, but its performance might suffer with extremely noisy or poorly aligned RGB-D data. In contrast, CMX, a transformer-based framework, optimizes RGB-X semantic segmentation. It utilizes two streams for feature extraction, applies a Cross-Modal Feature Rectification Module (CM-FRM) for spatial and channel-wise feature calibration, and uses a Feature Fusion Module (FFM) with a cross-attention mechanism for the final semantic prediction.

Despite the efficacy of the CMX model, its application of the Cross-Modal Feature Rectification Module (CM-FRM) and Feature Fusion Module (FFM) lacks resolution adaptability, and its dependence on max and average pooling operations overlooks the advantages of atrous convolutions. These offer superior spatial-temporal relationship maintenance and multi-scale feature leverage. We show that, integrating weighted atrous convolutions into the Cross-Attention based Encoder-Decoder Segmentation model boosts fine-detail capture across varied resolutions. Notably, previous methods chiefly use either CNNs or attention mechanisms, ignoring opportunities to optimize U-Net's long-range dependencies. Moreover, they may not be suitable for instance segmentation, particularly in robotic grasping applications where environmental dynamics complicate accurate object contour identification and subclass-level segmentation. Current methods also show insufficient generality in robotic grasping applications within industrial settings \cite{Hua2023DynamicTransmission, Karthik2022Contour-enhancedSegmentation, Wang2021EFNet:Segmentation}.


\section{{Bimodal SegNet}}
\label{section : Methodology}

\subsection{Prerequisite}
\subsubsection{\textbf{Event Signal}}
Vision cameras based on event-driven technology are designed to register variations in logarithmic light intensities by recording individual pixel-level alterations, referred to as events. These events generate a continuous event stream, which can be mathematically articulated as a series of ordered tuples. A tuple corresponds to an event $i$ and includes the event's spatial coordinates $(x_i, y_i)$, temporal stamps $t_i$, and polarity value $z_i$ \cite{BaghaeiNaeini2020AApplications}: 
    
\begin{equation}
    {(x_1, y_1, t_1, z_1), (x_2, y_2, t_2, z_2), ..., (x_n, y_n, t_n, z_n)}
\end{equation}

\subsubsection{\textbf{RGB Signal}}

Standard cameras capture visual data in the form of Red-Green-Blue (RGB) frames, each ranging from 0 to 255, which together represent the color of each individual pixel in the image. Mathematically, the color of a pixel located at $(x, y)$ in an RGB frame can be represented as a tuple $(R_{x,y}, G_{x,y}, B_{x,y})$. The entire RGB frame can therefore be represented as a matrix of these triplets across the spatial dimensions of the image.


\subsubsection{\textbf{Atrous convolution}}
Atrous convolution, also called dilated convolution, extends the receptive field in Convolutional Neural Networks (CNNs) without increasing computational load or parameters. It can be mathematically expressed as follows:

\begin{equation}
    \text{DilatedConv}\ {r_n}  = A[{x_i},{y_i}] = \sum_{k=1}^K C[i+r.k] \text{Conv}[k]
    \label{eqn: atrs}
\end{equation}

where $A[{x_i},{y_i}]$ is the 2D output feature map, $[{x_i},{y_i}]$ represents the location and $C$ is the input feature map. The convolution filter is stated by $Conv$ and the atrous stride rate is determined by $r$. The atrous rate $r$ increases the size of the kernel by inserting $r-1$ zeros along every spatial dimension. The filter’s receptive field is changed by modifying the stride rate \cite{Jun2022ACAU-Net:Segmentation}. In the case of standard convolution, the value of rate is $r = 1$.

\subsubsection{\textbf{Cross-Attention Mechanism}}

Cross-attention, a salient element of Transformer architectures, assigns weights to different elements within an alternate sequence, commonly originating from an encoder. Given an input vector sequence $D$ and a separate context modality sequence $H$, distinct learned linear transformations are utilized to derive Query $Q$ from $D$, and Key $K$ and Value $V$ from $H$. Each query-key score, $I$, is computed as the dot product of $Q$ and $K$, then scaled by the square root of the dimension of the key vector $d_k$. The m and n are indexing different elements in the input sequence. The attention scores (I) are calculated:

\begin{equation}
    I_{mn} = \frac{{Q_m K_n^T}}{{\sqrt{d_k}}}
\end{equation}

In essence, cross-attention permits each component in D to be influenced by elements from the context sequence H, in line with their contextual significance.

\subsection{\textbf{Bimodal SegNet}}

In this section, the proposed Bimodal SegNet architecture for robust instance segmentation in robotic grasping applications is presented in detail. The first section \ref{subsubsection : Event Synchronisation} explains the process of asynchronous event denoising and subsequent conversion into event frames. Next section \ref{subsubsection : Integrated Multi-sensory Encoder}  describes the Integrated Multisensory Encoder to handle Bimodal signals. Subsequently, Cross-Domain Contextual Attention, which facilitates the fusion of event and RGB features is discussed in section \ref{subsubsection : Cross-Domain Contexual Attention}. Further, the atrous Pyramidal Feature Amplification module is explained in section \ref{subsubsection : Atrous Pyramidal Feature Amplification}. Then, the concatenated skip connections between encoders and decoders are discussed in section \ref{subsubsection : Decoder}. Finally, the overall network architecture for the Bimodal SegNet is presented in section \ref{subsubsection : Proposed Network Architecture}.

\subsubsection{\textbf{Event Synchronisation}}
\label{subsubsection : Event Synchronisation}

A camera such as DAVIS can produce both continuous asynchronous events  (with temporal resolution of a few microseconds) and RGB frames (with frame rates between 25-50 Hz) simultaneously with consistent frame dimensions.
The configuration of the time window $T$ hyperparameter should consider parameters such as the DVS threshold, application speed, and noise, but most importantly the RGB frame rate. The events produced between two consecutive RGB frames are accumulated over a time window $T$ irrespective of position information \cite{Naeini2020Dynamic-vision-basedNetworks}. The event-to-frame conversion process can be formulated as:

\begin{equation}
    E_t(x,y,p) = \sum_{\forall e} rect( \frac{t_e}{T} - 0.5 - t) \delta_{xx_e} \delta_{yy_e} \delta_{pp_e}
    \label{eqn: event}
\end{equation}

where $E_t$  represents a frame of accumulated events for different polarities $p\in \{0,  1\}$ at timestamp \textit{t}. The Kronecker delta function and rectangle function are denoted as $\delta$ and \textit{rect} respectively. The $t_e$ represents the timestamp of each event and $e$ indicates the event number.

\subsubsection{\textbf{Integrated Multisensory Encoder}}
\label{subsubsection : Integrated Multi-sensory Encoder}

The architecture employs two separate encoders, akin to the contraction phase of the U-Net \cite{Lv2020AttentionSegmentation}, that operate concurrently on event frames and RGB frames. Each encoder follows the same pattern: it consists of four stages, each stage housing two 3x3 convolutions succeeded by ReLU activations and a 2x2 max pooling operation. The feature channels double at each stage, enabling the extraction of increasingly complex characteristics from the input frames.

Starting with the event and RGB frames of size 296x296, both frame types traverse their respective initial encoder stage. The padding of size 1 is used on all sides to maintain the image size while increasing the feature depth to 64. A max pooling operation subsequently halves the dimensions to 148x148. The frames then proceed through the following three stages within their respective encoders. The convolutions and max pooling operations repeat, cumulatively reducing spatial dimensions and increasing the feature depth. Post these operations, the dimensions successively transition to 148x148x128, 74x74x256, and finally 37x37x512 after the second, third, and fourth stages, respectively. Post the fourth stage, no max pooling operation is performed, producing two 37x37x512 feature maps - one each for the event and RGB frames. These feature maps produced at each stage of both encoders are fused using Cross-Domain Contextual Attention (CDCA), where they are merged to enhance spatial information. This fusion of feature maps from both event and RGB frames significantly contributes to improving the quality of the resultant segmentation.

\subsubsection{ \textbf{Cross-Domain Contexual Attention }}
\label{subsubsection : Cross-Domain Contexual Attention}

The Cross-Domain Contexual Attention (CDCA) module (Fig. \ref{fig:Cross Domain Contextual Attention}) is a pivotal component of the proposed Bimodal SegNet model. It is responsible for guiding the model's focus towards distinct portions of the complementary event signal. Let us denote the input feature maps from the RGB and event encoders as $F_{RGB}$ and $F_{event}$ respectively.

Initially, the input feature of dimensions $R^{H\times W\times C}$ is flattened to $R^{N\times C}$, where $N = H \times W$. Subsequently, a linear embedding is employed to generate two vectors of identical size $R^{N\times C}$, termed as the residual vectors $F_{RGB\_res}$, $F_{event\_res}$ and two interactive vectors $F_{RGB\_inter}$, $F_{event\_inter}$. An efficient cross-attention mechanism is then applied to these interactive vectors, further enhancing the information exchange process as follows:

\begin{figure*}[t!]
\resizebox{\hsize}{!}{\includegraphics[clip=true]{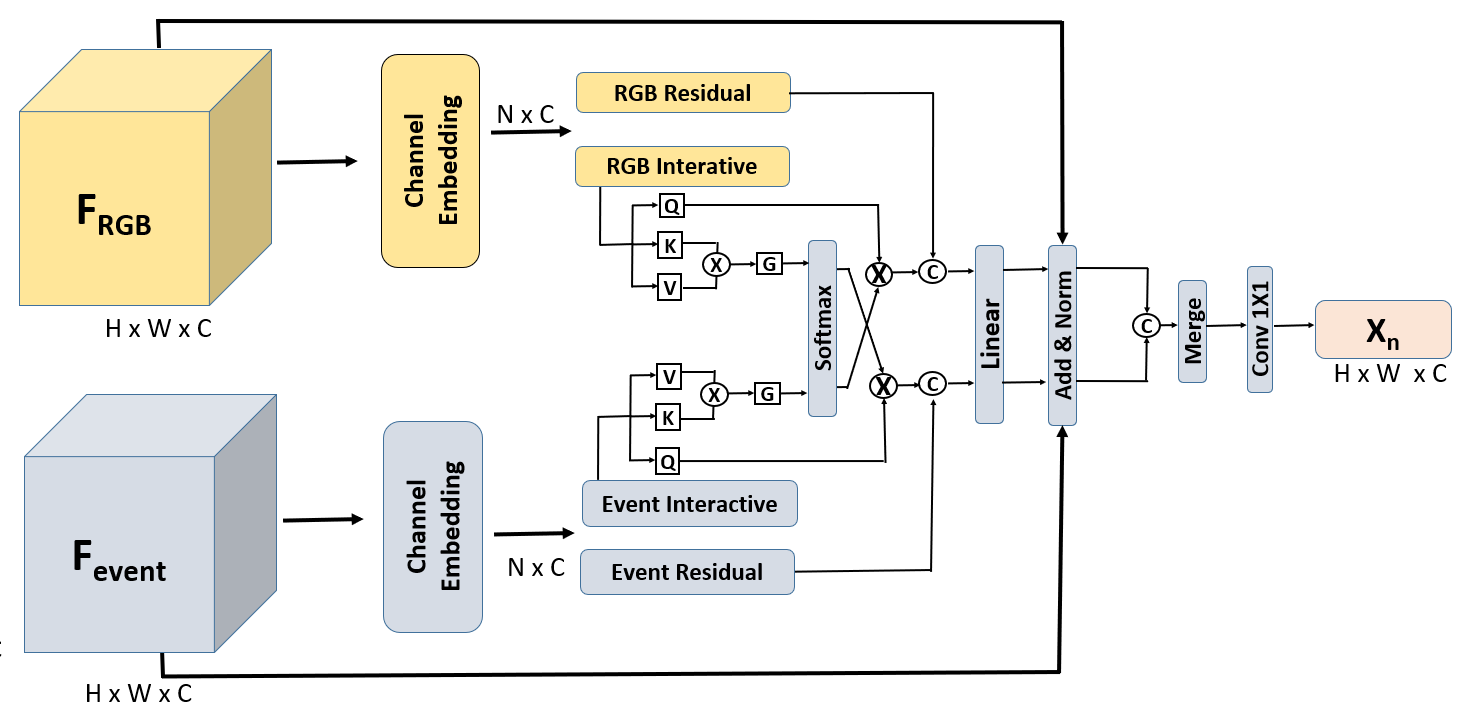}}
\caption{\footnotesize
{ Cross Domain Contextual Attention - The Bimodal SegNet utilizes Cross-Domain Contextual Attention (CDCA) to guide its focus on distinct portions of the signal, using input features from RGB and event encoders. The model employs an attention mechanism between the linearly embedded residual and interactive vectors. With each modality generating Query, Key, and Value matrices, a cross-attention process forms attended results. These results, along with the residual vectors, are then concatenated to create attention-augmented features, which combine with original features to produce enhanced representations. The final output is a fusion of information from both modalities, passed through a 1x1 convolution for consolidated, fused features.   }
}
\label{fig:Cross Domain Contextual Attention}
\end{figure*}

For each modality, the two corresponding interactive feature vectors $F_{RGB\_inter}$ and $F_{event\_inter}$ derived from the encoders undergo transformation to generate a set of Query (Q), Key (K), and Value (V) matrices. For the RGB modality, these transformations are denoted as:

\begin{equation}
    Q_{RGB} = W_{Q_{RGB}} \cdot F_{RGB\_inter}
\end{equation}

\begin{equation}
    K_{RGB} = W_{K_{RGB}} \cdot F_{RGB\_inter}
\end{equation}

\begin{equation}
    V_{RGB} = W_{V_{RGB}} \cdot F_{RGB\_inter}
\end{equation}

Conversely, for the event modality, the transformations are denoted as:

\begin{equation}
    Q_{event} = W_{Q_{event}} \cdot F_{event\_inter}
\end{equation}

\begin{equation}
    K_{event} = W_{K_{event}} \cdot F_{event\_inter}
\end{equation}

\begin{equation}
    V_{event} = W_{V_{event}} \cdot F_{event\_inter}
\end{equation}

where $W$ matrices represent trainable weights.

Specifically, the interactive $F_{inter}$ vectors are embedded into the corresponding $K$ and $V$ matrices for each modality, both possessing the dimensions $R_{N \times C_{head}}$. Global context vectors $G_{RGB}$ and $G_{event}$ are derived from $K$ and $V$ as follows: 

\begin{equation}
    G_{RGB} = K_{RGB}^{T} \times V_{RGB}
\end{equation}

\begin{equation}
    G_{event} = K_{event}^{T} \times V_{event}
\end{equation}

The $G_{RGB}$ and $G_{event}$ are further multiplied with the alternate modality path.  We refer to this operation as the cross-attention process, and is represented mathematically as follows:



\begin{equation}
    U_{RGB} = Q_{RGB} \times softmax\left(\frac{G_{event}}{\sqrt{d_k}}\right)
\end{equation}

\begin{equation}
    U_{event} = Q_{event} \times softmax\left(\frac{G_{RGB}}{\sqrt{d_k}}\right)
\end{equation}

Here, $U$ denotes the attended result and $d_k$ is the dimension of key vectors. The division by $\sqrt{d_k}$ acts as a scaling factor, mitigating the potential for excessively large values within the softmax function. To facilitate attention from diverse representation subspaces we retain the multi-head mechanism.

Subsequently, the attended result vectors $U$ and the residual vectors $X_{res}$ are concatenated as follows. Cross-attention augmented features are computed via the softmax function, evaluating the importance of distinct regions within the alternate modality:

\begin{equation}
    O_{RGB} = U_{RGB} \vert\vert F_{RGB\_res} 
\end{equation}

\begin{equation}
    O_{event} = U_{event} \vert\vert F_{event\_res}
\end{equation}

Further, a second linear embedding is applied, resizing the feature to $R_{H \times W \times C}$. Finally, the original feature maps are augmented with attention-augmented features to produce the ultimately enhanced representations:

\begin{equation}
    F_{RGB}^{'} = F_{RGB} + O_{RGB}
\end{equation}

\begin{equation}
   F_{event}^{'} = F_{event} + O_{event}
\end{equation}

These refined representations, $F_{RGB}^{'}$ and $F_{event}^{'}$, encapsulate the fused information from both modalities. The merged representation further passed into 1x1 convolution to change the size of the fused features $X_{n}$ to the size HxWxC.

\subsubsection{\textbf{Atrous Pyramidal Feature Amplification}}
\label{subsubsection : Atrous Pyramidal Feature Amplification}

In the proposed Bimodal SegNet architecture, the Atrous Pyramidal Feature Amplification (APFA) module is integrated at the end of two parallel encoders - an event encoder and an RGB encoder. The output of cross-domain enrichment mechanism $X_{n}$ at each downscaling stage is weighted and concatenated to pass into the APFA module. As the encoder captures increasingly abstract features at each of the 4 stages, given the model's objective to recover low-level features or contours, primarily from event representations, higher weights are assigned to the features extracted at later stages of the contraction path. The feature maps generated at each stage are combined in a weighted manner before being fed into the APFA module. Denoting an image as $I$ with dimensions height $H$ and width $W$, the CDCA outputs a series of feature maps $X_1, X_2, ..., X_n$ at each stage, each with dimensions $H' \times W' \times C$, where $C$ is the number of channels of the feature map. The weights for each feature map from stages one to four are $w_1$, $w_2$, $w_3$, and $w_4$. The empirical value of each weight has been discussed in experimental section \ref{subsection : Implementation details}. The weighted concatenation of these feature maps is denoted as $X_{\text{combined}}$:

\begin{equation}
X_{\text{combined}} = \text{Concat}(w_1 \cdot X_1, w_2 \cdot X_2, w_3 \cdot X_3, w_4 \cdot X_4)
\end{equation}

This combined feature map $F_{\text{combined}}$ is then processed through the APFA block. Given $N$ parallel branches in the APFA module with atrous rates $r_1, r_2, ..., r_N$, for each branch $n$, $X_{\text{combined}}$ is passed through a dilated convolution with the corresponding atrous rate, resulting in a new feature map $X_n$:

\begin{equation}
X_n = \text{DilatedConv}{r_n}(X{\text{combined}}) \quad \text{for} \quad n = 1, 2, ..., N.
\end{equation}

These feature maps are upsampled to the original spatial dimensions ( $H' \times W'$) as necessary and concatenated along the channel dimension to form a new combined feature map $X_{new_{combined}}$:

\begin{equation}
X_{new_{combined}} = \text{Concat}(X_1, X_2, ..., X_N)
\end{equation}

Finally, this combined feature map is passed through a 1x1 convolution to produce the final output $X_{\text{out}}$:

\begin{equation}
X_{out} = {Conv} (X_{new_{combined}})
\end{equation}





\subsubsection{\textbf{Decoder }}
\label{subsubsection : Decoder}

The decoder part of the Bimodal SegNet follows the architecture of the U-Net decoder, which is characterized by an expansion path that progressively upsamples the abstracted information back to the original image resolution \cite{Huang2022CM-UNet:Scenes}. The expansion path receives its input from the output of the APFA module, represented as $X_{\text{out}}$ in the previous section. As in the U-Net architecture, the expansion path in the Bimodal SegNet also includes skip connections from the encoder to the decoder. At each step of the expansion path, the feature maps from the corresponding stage of the encoder are concatenated with the upsampled output from the previous stage of the decoder. This operation helps to recover the high-resolution details lost during the encoding process. Suppose $X_{d, n}$ is the upsampled output from the $n$-th stage of the decoder, and $X_n$ is the feature map from the corresponding stage of the encoder. Then, the input to the $n+1$-th stage of the decoder, denoted as $X_{d, n+1}$, is given by:

\begin{equation}
    X_{d, n+1} = {Concat}(X_{d, n}, X_n).
\end{equation}

This operation is repeated at each stage of the decoder until the original image resolution is reached. The final output of the decoder, $X_{\text{seg}}$, represents the instance segmentation of the input image:

\begin{equation}
    X_{seg} = X_{d, N}.
\end{equation}

The use of such a decoder architecture enables the Bimodal SegNet to effectively combine the multi-scale context information captured by the ASPP module and the high-resolution details preserved by the encoder, resulting in high-quality instance segmentation results. While the intermediate layers of the U-Net architecture contain multiple feature maps, the final output layer is designed to produce a single feature map. This is achieved through a final 1x1 convolutional layer that maps the N+M feature maps down to the number of desired output classes. For multi-class segmentation tasks, a 1x1 convolution with C filters is applied (where C is the number of classes), followed by a softmax activation function, to output a C-channel image where each channel corresponds to the probability map of one class. The final segmentation is then usually obtained by assigning each pixel to the class with the highest probability at that pixel's location.

\begin{figure*}[h!]
\resizebox{\hsize}{!}{\includegraphics[clip=true]{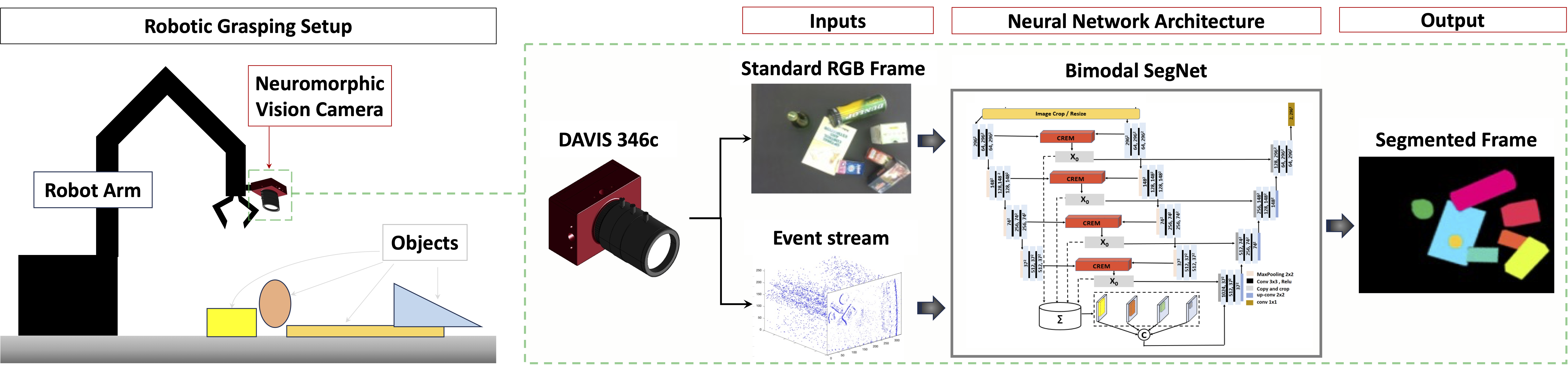}}
\caption{\footnotesize
{ The proposed approach, along with the Bimodal SegNet, strategically exploits dual modalities, thereby amalgamating the distinct strengths of each to enhance the efficacy of instance segmentation. This enhancement ensures robust object recognition even in challenging conditions such as occlusions, blurring, and variations in brightness, trajectory, and scale. Exemplary scenario: Eye-in-hand Robotic pick and place of cluttered objects in modern industries requiring quick operations under varying dynamic conditions. } }
\label{fig: Overview BimodalNet with arm}
\end{figure*}

\begin{figure*}[h!]
\resizebox{\hsize}{!}{\includegraphics[clip=true]{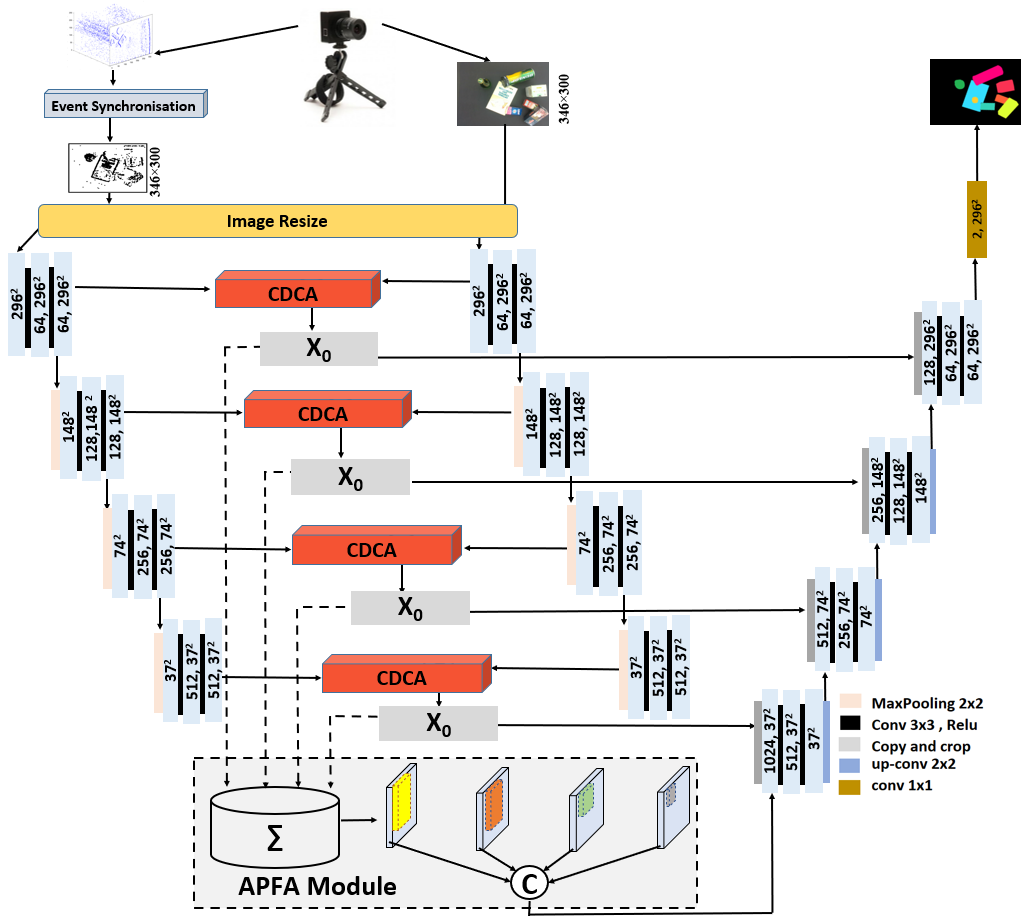}}
\caption{\footnotesize
{ The proposed Bimodal SegNet architecture uses event-based vision sensors such as DAVIS346 to produce both asynchronous events and RGB frames. These data are passed into Event Synchronisation and RGB encoders respectively. The convoluted blocks within these encoders downscale the input for multiple times to infer feature maps. At each downscaling stage, a CDCA layer is used, which inputs into the APFA block. The features for each sampling rate in the APFA block are then fused and sent to the decoder block. Here, the image is upscaled multiple times. The process uses a combination of up-convolution, copy and crop, and convolution with Relu, ultimately retrieving the original spatial dimension of the input image. The final fused tensor comes from the output of the CDCA module and the previous decoder layer. } }
\label{fig:BimodalNet}
\end{figure*}

\subsubsection{\textbf{Proposed Network Architecture}}
\label{subsubsection : Proposed Network Architecture}

To conduct the task of instance segmentation for robotic grasping, we utilize the event camera mounted on the robotic arm, as depicted in Fig. \ref{fig: Overview BimodalNet with arm}. Our approach involves performing instance segmentation of the objects in the scene using the Bimodal SegNet network architecture, detailed in Fig. \ref{fig:BimodalNet}.

Event-based vision sensors such as DAVIS346 can produce both continuous asynchronous events (with a temporal resolution of a few microseconds) and RGB frames (with frame rates between 25-50 Hz) simultaneously. The asynchronous events between two consecutive frames are first passed into the Event Synchronisation module where asynchronous events are converted to event frames. Both the event and RGB frames are then resized. The RGB frames are passed into RGB encoders $F_{RGB}$ and the event frames are passed into event encoders $F_{event}$. Both types of encoders are built with convolutional blocks for downscaling the input frame by 0.5x for $N+1$ consecutive times to learn or infer the feature maps. 

At each of the $N$ feature fusion stages in the encoder, the CDCA layer is used. The weighted feature maps from the CDCA layer at each downscaling stage are passed into the APFA block with different atrous rates. The features extracted for each sampling rate within the APFA block are fused at the output of the block. The fused results are then passed into the decoder block where the image is upscaled by 2x for $N$ times consecutively. At each of the $N$  feature fusion stages in the decoder, a combination of 2x2 up-convolution, copy and crop, 3x3 convolution with Relu is used. The up-convolution layer assists in retrieving the original spatial dimension of the input image. Further, the copy and crop layer takes the output from CDCA module i.e. $X_n$ and the output of the previous decoder layer to provide a single fused tensor.  


\section{{Experimentation}}
\label{section: Experimental Results}

\subsection{\textbf{Dataset}}

\begin{figure*}[h!]
    \centering
    \begin{tabular}{|c|c|c|c|c|c|c|}
    \hline
Type & Raw Image & RGB Label & Events Label     
\\

    \hline

    \rotatebox{90}{known} 
    \rotatebox{90}{2 objects}
    &\subfloat{\includegraphics[width = 0.8 in,height=0.6in]{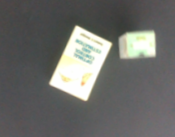}} 
    & \subfloat{\includegraphics[width = 0.8 in,height=0.6in]{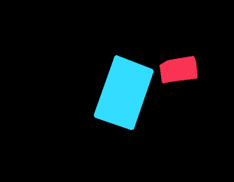}}
    & \subfloat{\includegraphics[width = 0.8 in,height=0.6in]{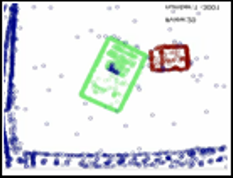}}
 \\
    \hline 

    \rotatebox{90}{known} 
   \rotatebox{90}{4 objects}
    &\subfloat{\includegraphics[width = 0.8 in,height=0.6in]{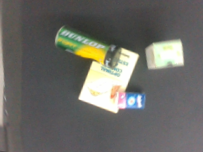}}
   & \subfloat{\includegraphics[width = 0.8 in,height=0.6in]{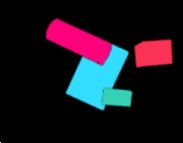}}
   & \subfloat{\includegraphics[width = 0.8 in,height=0.6in]{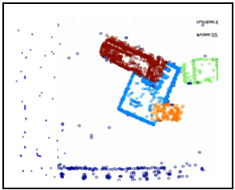}}
\\
    \hline
  
    \rotatebox{90}{known} 
    \rotatebox{90}{6 objects}
    &\subfloat{\includegraphics[width = 0.8 in,height=0.6in]{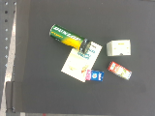}}
    &\subfloat{\includegraphics[width = 0.8 in,height=0.6in]{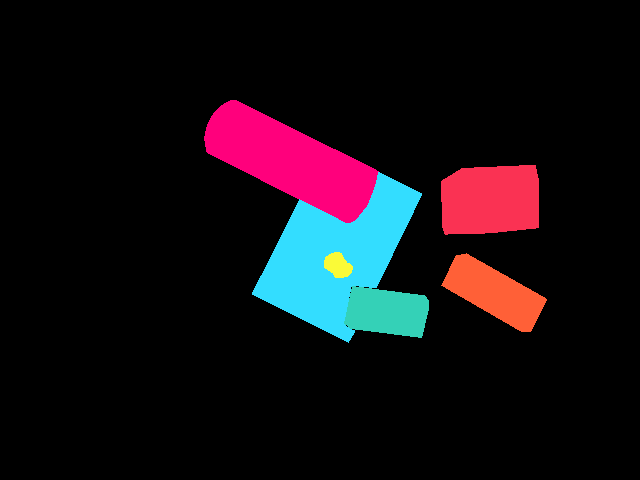}}
    &\subfloat{\includegraphics[width = 0.8 in,height=0.6in]{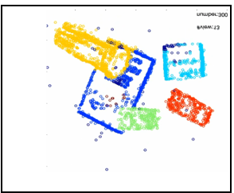}}
\\
    \hline
    
    \rotatebox{90}{known} 
    \rotatebox{90}{8 objects}
    &\subfloat{\includegraphics[width = 0.8 in,height=0.6in]{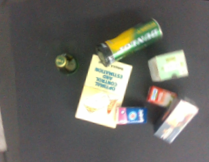}}
    &\subfloat{\includegraphics[width = 0.8 in,height=0.6in]{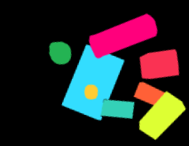}}
    &\subfloat{\includegraphics[width = 0.8 in,height=0.6in]{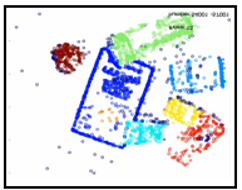}}
    \\
    \hline

    \rotatebox{90}{known} 
    \rotatebox{90}{10 objects}
    &\subfloat{\includegraphics[width = 0.8 in,height=0.6in]{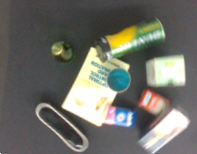}} 
    &\subfloat{\includegraphics[width = 0.8 in,height=0.6in]{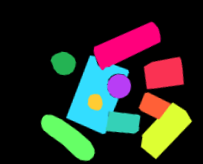}}
    &\subfloat{\includegraphics[width = 0.8 in,height=0.6in]{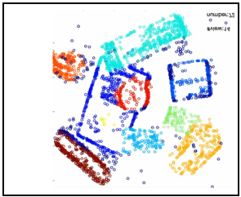}}
    \\
    \hline

    \rotatebox{90}{Unknown}  
    \rotatebox{90}{2 objects}
    &\subfloat{\includegraphics[width = 0.8 in,height=0.6 in]{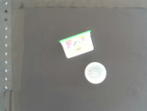}} 
    &\subfloat{\includegraphics[width = 0.8in,height=0.6in]{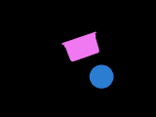}} 
    &\subfloat{\includegraphics[width = 0.8in,height=0.6in]{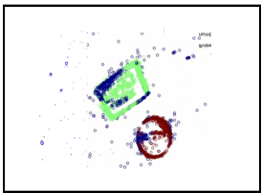}} 
    \\
    \hline

    \rotatebox{90}{Unknown}  
    \rotatebox{90}{5 objects}
    &\subfloat{\includegraphics[width = 0.8in,height=0.6in]{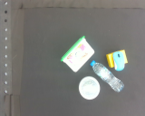}}
    &\subfloat{\includegraphics[width = 0.8in,height=0.6in]{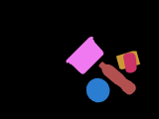}}
    &\subfloat{\includegraphics[width = 0.8in,height=0.6in]{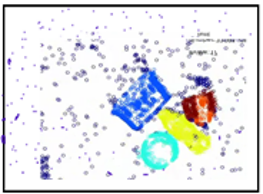}}\\

\hline

    \end{tabular}
    \caption{Example of the ESD-1 dataset (row 1-5) in terms of the number of known objects attributes, under the condition of 0.15 moving speed, normal light condition, linear movement, and 0.82 height. The ESD-2 dataset (rows 6,7) presents examples of previously unseen objects with varying attributes. Specifically, the dataset features scenes where objects are moving at a speed of 0.15, under normal lighting conditions, with linear motion, and at a height of 0.82. The RGB ground truth and annotated event mask use different colors to represent different object labels. For optimal understanding, it is recommended to view the dataset in color.}
    \label{fig: data_visualization_ESD}
 \end{figure*}

The ESD dataset, as described in \cite{Huang2023AEnvironment, huang2024neuromorphic}, is one of the largest datasets available for understanding instances of robotic grasping scenes. The dataset was captured using a DAVIS346 sensor mounted on a robotic arm and includes both conventional RGB frames and asynchronous events. The dataset also contains dense annotations for pixels and events, which are instance-specific and grouped into 6 categories, namely bottle, box, pouch, book, mouse, and platform. The dataset comprises 17186 annotated images and 177 labeled event streams. The ESD datast as shown in figure \ref{fig: data_visualization_ESD} has two sub-datasets, ESD-1 for known objects segmentation challenge and ESD-2 for an unknown object segmentation challenge. The ESD-1 dataset includes 10 objects depicted in both the training (13984 images) and the testing dataset (3202 images).  The ESD-2 dataset is only used for testing and includes five objects that differ from those in the ESD-1 dataset. Both datasets has variations in the direction of camera motion, arm speed, lighting conditions, and object clutter. The motion variations include linear, rotational, and partial-rotational motion, while the arm speed variants are 0.15 m/s, 0.3 m/s, and 1 m/s. The lighting conditions comprise normal light and low light. The number of objects in the clutter varies from 2 to 10, as shown in Fig. \ref{fig: data_visualization_ESD}. The RGB framerate is 40 Hz, therefore a temporal window of T=25 ms is used for framing the events to ensure temporal alignment between the two types of frames. 


\subsection{\textbf{Evaluation Metrics}}

\subsubsection{\textbf{Pixel accuracy}}

Pixel accuracy is a metric to calculate the percentage of pixels in the image that are classified correctly as given by Equation (\ref{eqn: Pixel Accuracy}) 

\begin{equation}
    Acc(p,p') = \frac{1}{N}\sum_{1}^N \delta (p_i,p'_i) \times 100
    \label{eqn: Pixel Accuracy}
\end{equation}

where $p$, $p'$, $N$ and $\delta$ represent the ground truth image, the predicted image, the total number of pixels and Kronecker delta function respectively. However, its descriptive power is limited in cases with a significant imbalance between foreground and background pixels.

\subsubsection{\textbf{Mean Intersection over Union (mIoU)}}

The performance of the network for instance segmentation is evaluated using the mean Intersection over Union (mIoU) also known as the Jaccard Index. mIoU handles imbalanced binary and multi-class segmentation. It is calculated across classes according to Equation (\ref{eqn: mIoU}):

\begin{equation}
    mIoU = \frac{1}{C}\sum_{i}^C \frac {\sum_{i}^N \delta (p_{i,c},1) \delta (p_{i,c},p'_{i,c})}{max (1,\delta (p_{i,c},1) + \delta (p'_{i,c},1) }
    \label{eqn: mIoU}
\end{equation}
 
Thus, the mIoU measure is more informative than the pixel accuracy as it considers false positives.

\subsection{\textbf{Implementation Details}}
\label{subsection : Implementation details}

The ResNet-101 pre-trained on ImageNet is the backbone of the model that is treated as an encoder. The last layer of the pre-trained ResNet-101 is replaced with 11 instances (10 objects and 1 background ). Both ESD-1 and ESD-2 datasets are divided into training/testing/validation with a ratio of 70:20:10. The DAVIS346 camera has a resolution of 346$\times$260, which differs from the network's expected input of 296$\times$296. Therefore, the image is first resized to 346$\times$300. Subsequently, the popular approach of patchifying the input images is employed and the final  segmentation map is reconstructed by unpatching the outputs. During training, data augmentation is performed by random flipping and scaling with random scales [0.5, 1.75]. Using empirical results \cite{Chen2018Encoder-DecoderSegmentation}, typically 6, 12, 18, 24 atrous rates are used. The weights for each feature map from stages one to four ie. $w_1$, $w_2$, $w_3$, $w_4$  are 10\%, 20\%, 30\%, and 40\% respectively. A 3$\times$3 kernel size for convolution is empirically selected as it is small enough to capture local spatial information. Softmax as an activation function and Adam as an optimizer are used while compiling the model. Categorical cross-entropy is used as a loss function. The fine-tuned model has 0.001 learning rate, 30 batch size, 100 epochs. The training was performed on a processor 11th Gen Intel(R) Core(TM) i7-1165G7, 1.69 GHz, 16 GB RAM and a Google Colab's NVIDIA Tesla K80 GPU.


\subsection{\textbf{ESD-1 Dataset - Experimental Results and Analysis}}
\label{section:Results ESD-1 Dataset}

To assess the robustness of the model, experiments were conducted, varying object clutter, lighting, camera distance, trajectory, and speed. The proposed model was compared to state-of-the-art models, including RGB-event fusion models RFNet\cite{Sun2020Real-TimeDetection}, ISSAFE \cite{Zhang2020ISSAFE:Data}, SA-Gate \cite{Xiaokang2020Bi-directionalSegmentation}, CMX \cite{Liu2022CMX:Transformers}. All the models used in this study are trained on the ESD-1 dataset \cite{Huang2023AEnvironment} using default parameters for feature fusion of multiple modalities described in \cite{Liu2022CMX:Transformers}. If not explicitly specified, the following default conditions will be considered to define the testing subsets for the experiments: 2 objects, normal light, rotational arm motion, 0.15 m/s speed and 62 cm platform to camera distance. Comparative results are presented in Table \ref{tab: ESD1}.

\begin{table}[t]
    \centering
    \caption{Segmentation accuracy of known objects in various conditions.}
  
   \begin{adjustbox}{width=0.5\textwidth}

    \centering
      {\fontsize{10}{12}\selectfont
    \begin{tabular}{lccccc}
  
  \specialrule{.15em}{.1em}{.1em}  
        \multicolumn{6}{c}{Exp-1: \textbf{varying clutter objects}, Bright light, 62cm height,  Rotational motion, 0.15 m/s speed } \\
  \specialrule{.1em}{.1em}{.1em} 
        Method & 2 Obj & 4 Obj & 6 Obj & 8 Obj & 10 Obj   \\  
  \specialrule{.1em}{.1em}{.1em}   

        RFNet\cite{Sun2020Real-TimeDetection} & 65\% &	64\%	& 62\%	& 61\% &	58\%     \\
        ISSAFE\cite{Zhang2020ISSAFE:Data} & 71\% &	69\% &	68\% &	65\% &	62\% \\
        SA-Gate\cite{Xiaokang2020Bi-directionalSegmentation} & 78\%	& 77\%	& 75\%	& 73\%	& 69\% \\
        CMX\cite{Liu2022CMX:Transformers} & 92\%	& 91\%	& 84\%	& 84\%	& 85\% \\
        Ours & \textbf{95\%}  & \textbf{95\%} & \textbf{92\%} &  \textbf{89\%} &  \textbf{87\%} \\
   \specialrule{.15em}{.1em}{.1em}

   \specialrule{.15em}{.1em}{.1em}  
        \multicolumn{6}{c}{Exp-2: 6 Objects, \textbf{varying lighting conditions}, 62cm height, Rotational Motion, 0.15 m/s speed.} \\
  \specialrule{.1em}{.1em}{.1em} 
        Method & Bright Light & Low light &  &  &    \\  
    \specialrule{.1em}{.1em}{.1em}

        RFNet\cite{Sun2020Real-TimeDetection} & 58\%	& 53\%   &  &  &      \\
        ISSAFE\cite{Zhang2020ISSAFE:Data} &62\%	& 55\%  &    &   &   \\
        SA-Gate\cite{Xiaokang2020Bi-directionalSegmentation} & 81\%	& 76\%  &   &    &   \\
        CMX\cite{Liu2022CMX:Transformers} & 87\%	& 82\%  &   &    &   \\
        Ours & \textbf{89\%}  & \textbf{86\%} &  &   &   \\  
   \specialrule{.15em}{.1em}{.1em}

   \specialrule{.15em}{.1em}{.1em}  
        \multicolumn{6}{c}{Exp-3: 6 Objects, Bright Light, 62cm height, \textbf{Varying directions of motion}, 0.15 m/s speed.} \\
  \specialrule{.1em}{.1em}{.1em} 
        Method & Linear & Rotational & Partial Rotational &  &    \\  
\specialrule{.1em}{.1em}{.1em}

        RFNet\cite{Sun2020Real-TimeDetection} & 51\% &	58\% &	55\%  &  &      \\
        ISSAFE\cite{Zhang2020ISSAFE:Data} & 59\% &	70\% &	65\%   &   &   \\
        SA-Gate\cite{Xiaokang2020Bi-directionalSegmentation} & 72\% &	79\% &	76\%   &    &   \\
        CMX\cite{Liu2022CMX:Transformers} & 84\% &	89\% &	85\%   &    &   \\
        Ours & \textbf{85\%}  &  \textbf{91\%}  &  \textbf{87\%}  &   &   \\  
   \specialrule{.15em}{.1em}{.1em}

   \specialrule{.15em}{.1em}{.1em}  
        \multicolumn{6}{c}{Exp-4: 6 Objects, Bright Light, 62cm height, Rotational motion, \textbf{Varying speed}.} \\
  \specialrule{.1em}{.1em}{.1em} 
        Method & 0.1 m/s & 0.15 m/s &   0.3 m/s &  &    \\  
  \specialrule{.1em}{.1em}{.1em}

        RFNet\cite{Sun2020Real-TimeDetection} & 64\% &	61\% &	58\%  &  &      \\
        ISSAFE\cite{Zhang2020ISSAFE:Data} & 67\% &	65\% &	62\%   &   &   \\
        SA-Gate\cite{Xiaokang2020Bi-directionalSegmentation} &  75\% &	73\% &	71\%    &    &   \\
        CMX\cite{Liu2022CMX:Transformers} &  84\% &	79\% &	74\%    &    &   \\
        Ours & \textbf{89\% }  &  \textbf{88\%}  &  \textbf{85\%}  &   &   \\  
   \specialrule{.15em}{.1em}{.1em}

   \specialrule{.15em}{.1em}{.1em}  
        \multicolumn{6}{c}{Exp 5: 6 Objects, Bright Light, \textbf{Varying camera height}, Rotational motion, 0.15 m/s speed.} \\
  \specialrule{.1em}{.1em}{.1em} 
        Method & 62 cm & 82 cm &    &   &    \\  
  \specialrule{.1em}{.1em}{.1em}

        RFNet\cite{Sun2020Real-TimeDetection} & 68\% &	65\%  &   &  &      \\
        ISSAFE\cite{Zhang2020ISSAFE:Data} & 72\% &	71\%  &     &   &   \\
        SA-Gate\cite{Xiaokang2020Bi-directionalSegmentation} &  79\% &	76\%  &     &    &   \\
        CMX\cite{Liu2022CMX:Transformers} &  86\% &	85\%  &     &    &   \\
        Ours & \textbf{89\% }  &  \textbf{86\%}  &    &   &   \\  
   \specialrule{.15em}{.1em}{.1em}

   \specialrule{.15em}{.1em}{.1em}  
        \multicolumn{6}{c}{Exp 6: ESD-1 Validation Dataset} \\
  \specialrule{.1em}{.1em}{.1em}

        RFNet\cite{Sun2020Real-TimeDetection} & 60\% &	 &   &  &      \\
        ISSAFE\cite{Zhang2020ISSAFE:Data} & 65\% &	  &     &   &   \\
        SA-Gate\cite{Xiaokang2020Bi-directionalSegmentation} &  76\% &	  &     &    &   \\
        CMX\cite{Liu2022CMX:Transformers} &  86\% &	  &     &    &   \\
        Ours & \textbf{88\% }  &    &    &   &   \\  
   \specialrule{.15em}{.1em}{.1em}    
   \specialrule{.15em}{.1em}{.1em}

    \end{tabular}}
    \end{adjustbox}
    \label{tab: Segmentation Accuracy in Various conditions}
     \setlength{\belowcaptionskip}{-12pt}\label{tab: ESD1}
\end{table}

\subsubsection{\textbf{\textit{Exp-1:Varying clutter objects/Occlusion}}}
\label{subsubsection : pre-trained occlusion results}

The initial experiment was conducted on a sub-dataset wherein the count of objects interspersed within clutter varies between 2, 4, 6, 8, and 10. The scenario of two objects denotes an absence of occlusion, while scenarios with an object count exceeding two denote the presence of occlusions. Moreover, a commensurate increase in occlusion and scenario complexity is observed with the increase in the number of objects. This pattern is endorsed by the segmentation accuracy outcomes across all models. As the quantity of objects rises, a proportional decrease is observed in the segmentation accuracy of all models, which is gauged here by the metric of mean Intersection over Union (mIoU). For example, with a pair of objects, the mIoU scores for all models span from 65\% (RFNet) to 92\% (CMX). In contrast, with ten objects in place, these scores plummet to a range of 58\% (RFNet) to 85\% (CMX). The performance of our proposed method ranges from  95\% mIoU for two objects, to 87\% for ten objects. Our method retains a superior performance for all object-related scenarios in comparison to well-established models such as RFNet, ISSAFE, SA-Gate, and CMX, thereby underlining its efficacy in handling scenarios with varying degrees of clutter and occlusion.

\subsubsection{\textbf{\textit{Exp-2: Varying lighting conditions}}} 
\label{subsubsection : light conditions results}

In the experimental investigation carried out on Sub-dataset 2, the variables of luminosity, specifically the contrast between high and low illumination, were modified, while the number of objects, camera height, nature of motion, and speed were kept constant. All models evidenced a decrease in accuracy for low lighting conditions, indicating the inherent challenges in maintaining performance in such circumstances. Nevertheless, there is a consistent reduction in accuracy, roughly within the range of 5-7\% across all competitor models, illustrating the uniform sensitivities of the different models to alterations in lighting. For example, the RFNet, ISSAFE, SA-Gate, and CMX models experienced a notable reduction in accuracy between 5\%-7\%. Our model, however, showcased exceptional resilience with a marginal decrease of only 3\%, from an initial 89\% to 86\%. These findings corroborate the robustness of the proposed model under diverse lighting conditions.

\subsubsection{\textbf{\textit{Exp-3: Varying moving trajectories}}} 
\label{subsubsection : Trajecotries results}

In the ensuing experiment, executed on a subset of the test dataset, the direction of robotic arm movement was manipulated to follow linear, rotational, or partially rotational paths. With the use of event-based vision sensors, it was observed that the orientation of the robotic arm's movement significantly influenced the outcomes, as perpendicular trajectories produced more informative event sets compared to parallel ones. Linear motion, characterized by abrupt directional shifts, resulted in higher jerk and vibrations, contributing to a higher prevalence of blurred images compared to rotational and partially rotational motion, which involved smoother trajectories. This impact was explicitly illustrated in subset for this experiment, which integrated variations in the movement of objects, categorized as linear, rotational, and partial rotational. The experimental results demonstrated the highest segmentation accuracy of 85\% under linear motion for the proposed model, followed by CMX at 84\%, SA-Gate at 72\%, ISSAFE at 59\%, and RFNet at 51\%. Highest performance patterns were observed under rotational motion, with our model achieving an impressive accuracy of 91\%, followed by CMX at 89\%, SA-Gate at 79\%, ISSAFE at 70\%, and RFNet at 58\%, supporting the fact that higher events are triggered in rotational motion than linear motion. For partial rotational motion, the proposed model maintained its leading position with an accuracy of 87\%. This consistent performance supremacy of the proposed model across varied types of motion underscores its effective capability in managing diverse movement trajectories.

\subsubsection{\textbf{\textit{Exp-4: Moving speed of camera}}}
\label{subsubsection : speed results}
Experiment 4 scrutinizes the models' efficacy under diverse motion speeds, fluctuating between 0.15 m/s, 0.3 m/s, and 0.1 m/s. Positioned at the extremity of the robotic arm, the camera's speed and consequent image blur becomes a critical determinant when assessing the robustness of the model. It is notable that as the speed escalated from 0.1 m/s to 0.3 m/s, there was a universal decrement in the segmentation accuracy across all models. Nevertheless, our proposed model demonstrated the lowest reduction in accuracy, descending from 89\% at 0.1 m/s to 85\% at 0.3 m/s. In contrast, CMX experienced the highest decline in performance, from  84\% to 74\%. In spite of the fluctuations in speed, our model consistently showcased the highest accuracy, attesting to its steadfast performance under diverse motion speeds. The significance of event-based vision becomes especially evident under high-speed conditions, where it facilitates precise contour estimation, thereby substantiating its essential role in maintaining the accuracy of the model.

\subsubsection{\textbf{\textit{Exp-5: Object size variance}}} 
\label{subsubsection : size variance results}

To ascertain the model's scale invariance, an experiment was conducted on a subset of the training dataset wherein the distance between the camera and the platform was alternated between 62 cm and 82 cm. It should be noted that an increment in camera height typically induces transformations in the object's perspective and scale, consequently influencing the segmentation accuracy. A decrease in accuracy at the escalated camera height was observed across all models, with RFNet demonstrating a reduction from 68\% to 65\%, ISSAFE from 72\% to 71\%, SA-Gate from 79\% to 76\%, and CMX from 86\% to 85\%. Our proposed model maintained superior performance at both heights with a decline in accuracy from 89\% to 86\%. This observation reinforces the model's robustness against changes in camera perspective and object scale, highlighting the crucial role of event-based vision in maintaining accuracy in the face of object size variance.

\subsubsection{\textbf{\textit{Exp-6: Whole Dataset}}} 
\label{subsubsection : Whole Dataset}

Finally, the comprehensive results derived from the 'Whole Dataset' illustrate an overarching evaluation of all models under a diversity of conditions. The proposed model recorded an impressive accuracy of 88\%, surpassing the performance of all competing models. This not only confirms its superior performance but also accentuates its adaptability and robustness across a wide range of conditions and parameters. Importantly, the results underscore the significant role of the event-based vision camera in architecture. The model's effectiveness can be attributed to the combination of the Cross-Domain Context Awareness (CDCA) module and the Atrous Pyramidal Feature Attention (APFA) module, both integral to the architecture. The CDCA module, employing a cross-attention mechanism, expertly manages disparity across domains while the APFA module, employing weighted atrous pyramidal pooling, significantly enhances contour feature extraction, therefore contributing to the model's superior accuracy and resilience to varying conditions.

\subsection{\textbf{ESD-2 Dataset - Experimental Results and Analysis}}
\label{section:Results ESD-2 Dataset}

\textbf{ESD-2 Dataset :} All the models used in this study are trained on ESD-1 , but tested on the ESD-2 \cite{Huang2023AEnvironment}, with no common objects between the two datasets. The proposed model was compared to state-of-the-art RGB-event fusion models RFNet, ISSAFE, SA-Gate, CMX\cite{Liu2022CMX:Transformers}. If not explicitly specified, the following
default conditions will be considered to define the testing subsets for the experiments: 2 objects, normal light, rotational arm motion, 0.15 m/s speed and 62 cm platform to camera distance.  Comparative results are presented in Table \ref{tab: ESD2}.

\begin{table}[t]
    \centering
    \caption{Segmentation accuracy of known objects in various conditions.}
    
   \begin{adjustbox}{width=0.5\textwidth}

    \centering
      {\fontsize{10}{12}\selectfont
    \begin{tabular}{lccccc}
  
  \specialrule{.15em}{.1em}{.1em}  
        \multicolumn{6}{c}{Exp 1: \textbf{varying clutter objects}, Bright light, 62cm height,  Rotational motion, 0.15 m/s speed } \\
  \specialrule{.1em}{.1em}{.1em} 
        Method & 2 Obj & 5 Obj &  &  &    \\  
  \specialrule{.1em}{.1em}{.1em}

        RFNet\cite{Sun2020Real-TimeDetection} & 57\% &	55\%	& 	&  &	     \\
        ISSAFE\cite{Zhang2020ISSAFE:Data} & 60\% &	59\% &	& & \\
        SA-Gate\cite{Xiaokang2020Bi-directionalSegmentation} & 69\% &	67\%	&	& & \\
        CMX\cite{Liu2022CMX:Transformers} & 81\% &	79\%	&	& & \\
        Ours & \textbf{86\%}  & \textbf{84\%} &  &   &   \\
   \specialrule{.15em}{.1em}{.1em}

   \specialrule{.15em}{.1em}{.1em}  
        \multicolumn{6}{c}{Exp 2: 6 Objects, \textbf{varying lighting conditions}, 62cm height, Rotational Motion, 0.15 m/s speed.} \\
  \specialrule{.1em}{.1em}{.1em} 
        Method & Bright Light & Low light &  &  &    \\  
\specialrule{.1em}{.1em}{.1em}

        RFNet\cite{Sun2020Real-TimeDetection} & 49\% &	42\%   &  &  &      \\
        ISSAFE\cite{Zhang2020ISSAFE:Data} & 54\% &	46\%  &    &   &   \\
        SA-Gate\cite{Xiaokang2020Bi-directionalSegmentation} & 70\% &	65\%  &   &    &   \\
        CMX\cite{Liu2022CMX:Transformers} & 77\% &	72\%  &   &    &   \\
        Ours & \textbf{79\%}  & \textbf{75\%} &  &   &   \\  
   \specialrule{.15em}{.1em}{.1em}

   \specialrule{.15em}{.1em}{.1em}  
        \multicolumn{6}{c}{Exp 3: 6 Objects, Bright Light, 62cm height, \textbf{Varying directions of motion}, 0.15 m/s speed.} \\
  \specialrule{.1em}{.1em}{.1em} 
        Method & Linear & Rotational & Partial Rotational &  &    \\  
\specialrule{.1em}{.1em}{.1em}
  
        RFNet\cite{Sun2020Real-TimeDetection} & 40\% &	48\% &	44\%  &  &      \\
        ISSAFE\cite{Zhang2020ISSAFE:Data} & 48\% &	60\% &	53\%   &   &   \\
        SA-Gate\cite{Xiaokang2020Bi-directionalSegmentation} & 62\% &	68\% &	65\%   &    &   \\
        CMX\cite{Liu2022CMX:Transformers} & 73\% &	77\% &	74\%   &    &   \\
        Ours & \textbf{75\%}  &  \textbf{80\%}  &  \textbf{76\%}  &   &   \\  
   \specialrule{.15em}{.1em}{.1em}

   \specialrule{.15em}{.1em}{.1em}  
        \multicolumn{6}{c}{Exp 4: 6 Objects, Bright Light, 62cm height, Rotational motion, \textbf{Varying speed}.} \\
  \specialrule{.1em}{.1em}{.1em} 
        Method & 0.1 m/s & 0.15 m/s &   0.3 m/s &  &    \\  
  \specialrule{.1em}{.1em}{.1em}

        RFNet\cite{Sun2020Real-TimeDetection} & 53\% &	50\% &	47\%  &  &      \\
        ISSAFE\cite{Zhang2020ISSAFE:Data} & 57\% &	55\% &	52\%   &   &   \\
        SA-Gate\cite{Xiaokang2020Bi-directionalSegmentation} &  65\% &	63\% &	61\%    &    &   \\
        CMX\cite{Liu2022CMX:Transformers} &  76\% &	76\% &	74\%    &    &   \\
        Ours & \textbf{79\% }  &  \textbf{78\%}  &  \textbf{74\%}  &   &   \\  
   \specialrule{.15em}{.1em}{.1em}

   \specialrule{.15em}{.1em}{.1em}  
        \multicolumn{6}{c}{Exp 5: 6 Objects, Bright Light, \textbf{Varying camera height}, Rotational motion, Varying speed.} \\
  \specialrule{.1em}{.1em}{.1em} 
        Method & 62 cm & 82 cm &    &   &    \\  
  \specialrule{.1em}{.1em}{.1em}

        RFNet\cite{Sun2020Real-TimeDetection} & 59\% &	55\%  &   &  &      \\
        ISSAFE\cite{Zhang2020ISSAFE:Data} & 72\% &	71\%  &     &   &   \\
        SA-Gate\cite{Xiaokang2020Bi-directionalSegmentation} &  68\% &	66\%  &     &    &   \\
        CMX\cite{Liu2022CMX:Transformers} & 75\% &	74\%  &     &    &   \\
        Ours & \textbf{79\% }  &  \textbf{76\%}  &    &   &   \\  
   \specialrule{.15em}{.1em}{.1em}

   \specialrule{.15em}{.1em}{.1em}  
        \multicolumn{6}{c}{Exp 6: ESD-1 Validation Dataset} \\
  \specialrule{.1em}{.1em}{.1em}

        RFNet\cite{Sun2020Real-TimeDetection} & 46\% &	 &   &  &      \\
        ISSAFE\cite{Zhang2020ISSAFE:Data} & 56\% &	  &     &   &   \\
        SA-Gate\cite{Xiaokang2020Bi-directionalSegmentation} &  63\% &	  &     &    &   \\
        CMX\cite{Liu2022CMX:Transformers} &  73\% &	  &     &    &   \\
        Ours & \textbf{79\% }  &    &    &   &   \\  
   \specialrule{.15em}{.1em}{.1em}    
   \specialrule{.15em}{.1em}{.1em}

    \end{tabular}}
    \end{adjustbox}
    \label{tab: Segmentation Accuracy in Various conditions}
     \setlength{\belowcaptionskip}{-12pt}
     \label{tab: ESD2}
\end{table}

\subsubsection{\textbf{\textit{Exp-1:varying clutter objects/Occlusion}}}
\label{subsubsection : pre-trained occlusion results}

In Experiment 1, the segmentation accuracy of unknown objects was evaluated under conditions with varying levels of clutter. Clutter refers to the number of objects present within the scene, with two and five objects being the conditions examined in this study. The results showed that all models experienced a decline in accuracy as the number of unknown objects increased. Nevertheless, our model consistently outperformed the other methods, maintaining an accuracy of 86\% with two objects and 84\% with five objects, demonstrating a high level of generalization ability.

\subsubsection{\textbf{\textit{Exp-2: Varying lighting condition}}} 
\label{subsubsection : light conditions results}

Experiment 2 analyzed the performance of the models under varying lighting conditions, which were bright light and low light. Just as with known objects, the accuracy of all models decreased under lower lighting conditions when dealing with unknown objects. However, our model demonstrated superior adaptability, achieving the highest accuracy in both bright (79\%) and low light (75\%) conditions. This underlines the model's ability to handle unknown objects under different lighting conditions effectively.

\subsubsection{\textbf{\textit{Exp-3: Varying moving trajectories}}} 
\label{subsubsection : Trajecotries results}

In Experiment 3, the accuracy of the segmentation of unknown objects was evaluated under different directions of motion: linear, rotational, and partial rotational. Despite the changing motion directions, our model demonstrated superior performance with accuracy rates of 75\% for linear, 80\% for rotational, and 76\% for partial rotational motion. These results reflect the model's effectiveness in segmenting unknown objects under various motion trajectories, which is an essential quality for dynamic environments.

\subsubsection{\textbf{\textit{Exp-4: Moving speed of camera}}}
\label{subsubsection : speed results}

Experiment 4 focused on the effect of varying speeds on the models' performance. As the speed increased from 0.1 m/s to 0.3 m/s, all models displayed a decrease in segmentation accuracy when dealing with unknown objects. However, our model was the most resilient to speed changes, maintaining the highest accuracy at all speed levels: 79\% at 0.1 m/s, 78\% at 0.15 m/s, and 74\% at 0.15 m/s. These findings underscore the model's capability in maintaining reliable performance even at higher speeds, which is vital for real-world, high-speed applications.

\subsubsection{\textbf{\textit{Exp-5: Object size variance}}} 
\label{subsubsection : size variance results}

Experiment 5 investigated the models' performance at different camera heights, specifically at 62 cm and 82 cm. All models showed a decrease in accuracy with the increase in camera height, indicative of the challenges in maintaining perspective and scale when dealing with unknown objects. Yet, our model proved to be the most robust, demonstrating a minimal decrease in accuracy from 79\% at 62 cm to 76\% at 82 cm.

\subsubsection{\textbf{\textit{Exp-6: ESD-2 Whole Dataset}}} 
\label{subsubsection : Whole Dataset}

Lastly, the 'Whole Dataset' results presented a comprehensive evaluation of the models under all conditions with unknown objects. Our model excelled with an impressive accuracy of 79\%, outperforming all the other models and showcasing its robustness and generalization ability across various conditions and parameters, which is pivotal for practical, real-world applications.

\subsection{\textbf{Efficiency Analysis}}

Table \ref{tab:Efficiency Results} presents a comparative analysis of our proposed method against several state-of-the-art techniques, in terms of model complexity, computational cost, and segmentation performance. The model complexity is measured by the number of parameters  (number of Parameters in millions), while the computational cost is evaluated in terms of floating-point operations (FLOPs in billions, G). Performance is evaluated on the ESD-1 dataset using the mean Intersection over Union (mIoU) metric, expressed in percentage. The methods compared are RFNet, ISSAFE, SA-Gate, CMX (Segformer-B5), and our proposed method. As shown, our approach outperforms all other methods in terms of mIoU, achieving a score of 83.2\%, indicating superior segmentation performance. Moreover, our method exhibits a good balance between model complexity and computational cost, highlighting its efficiency. The proposed method has the lowest computational cost, even when compared to models with much lower complexity as RFNet. At the same time, our model's complexity is more than 30\% lower than the main competitor's (CMX). It is important to note that the FLOP counts were estimated for inputs consisting of both RGB and Event data, each of size 300$\times$300$\times$3.

\begin{table}[ht]
\centering
\caption{Efficiency results. FLOPs are estimated for inputs of
RGB (300×300×3) and Event data (300×300×3).}
\begin{tabular}{|c|c|c|c|}
\hline
\textbf{Method} & \textbf{\#Params (M)} & \textbf{FLOPs (G)} & \textbf{mIoU (\%)} \\
\hline
RFNet\cite{Sun2020Real-TimeDetection}  & \textbf{61.7}   & 147.2   & 58.4 \\
\hline
ISSAFE\cite{Zhang2020ISSAFE:Data} & 79.8   &  163.1  & 66.2   \\
\hline
SA-Gate\cite{Xiaokang2020Bi-directionalSegmentation} & 63.4 & 204.9 & 69.3 \\
\hline
CMX (Segformer-B5) & 181.1 & 167.8 & 76.1 \\
\hline
Ours &  123.8   & \textbf{126.2}    &  \textbf{83.2}   \\
\hline
\end{tabular}

\label{tab:Efficiency Results}
\end{table}

\subsection{Ablation Study}
\subsection{Effectiveness of APFA and CDCA}
Table \ref{tab:ablation_study} demonstrates the effectiveness of the APFA and CDCA modules through an ablation study conducted on the ESD-1 test set. The first row presents the baseline, where neither APFA nor CDCA is applied, and features are merely averaged for semantic prediction, resulting in an mIoU of 66.9\% and pixel accuracy of 79.2\%. Introducing APFA (as depicted in the second row) significantly improves the model's performance, elevating mIoU to 75.3\% and pixel accuracy to 84.8\%, underscoring the importance of adaptive feature extraction. Employing CDCA without APFA (as shown in the third row) also brings a boost, albeit slightly less, with an mIoU of 70.1\% and pixel accuracy of 82.4\%. This indicates the importance of cross-modal attention in feature fusion and denoising for obtaining sharp object boundaries. However, the most impressive performance is achieved when APFA and CDCA are utilized concurrently, raising the mIoU to 83.2\% and pixel accuracy to an impressive 87.2\%. These improvements collectively highlight that both APFA and CDCA modules are integral for enhancing RGB-Event instance segmentation performance.

\begin{table}[ht]
\centering
\caption{Ablation study of APFA and CDCA modules on ESD-1 test set. Avg. is the average fusion.}
\begin{tabular}{|c|c|c|c|}
\hline
\textbf{APFA} & \textbf{CDCA} & \textbf{mIoU (\%)} & \textbf{Pixel Acc. (\%)} \\
\hline
$\times$ & Avg. &  66.9  & 79.2 \\
\hline
$\checkmark$ & Avg. & 75.3  & 84.8  \\
\hline
$\times$ & $\checkmark$ & 70.1  & 82.4  \\
\hline
$\checkmark$ & $\checkmark$ & \textbf{83.2}  & \textbf{87.2}  \\
\hline
\end{tabular}
\label{tab:ablation_study}
\end{table}

\subsection{Qualitative Results}

\begin{figure*}[h!]
\centering
\begin{adjustbox}{width=0.8\textwidth}
\begin{tabular}{|c|c|c|c|c|c|c|c|c|c|c|}
\hline
Sample & Raw Image & RGB GT & Event GT & CMX  & Ours  \\

\hline  

\rotatebox{90}{ESD-1 :Exp-3}
\rotatebox{90}{Rotational Motion }
&\subfloat{\includegraphics[width = 1 in,height= 1 in]{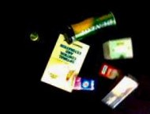}}  
& \subfloat{\includegraphics[width = 1 in,height= 1 in]{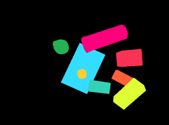}}
& \subfloat{\includegraphics[width = 1 in,height= 1 in]{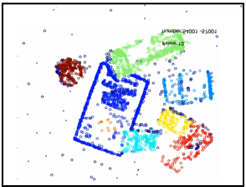}}
& \subfloat{\includegraphics[width = 1 in,height= 1 in]{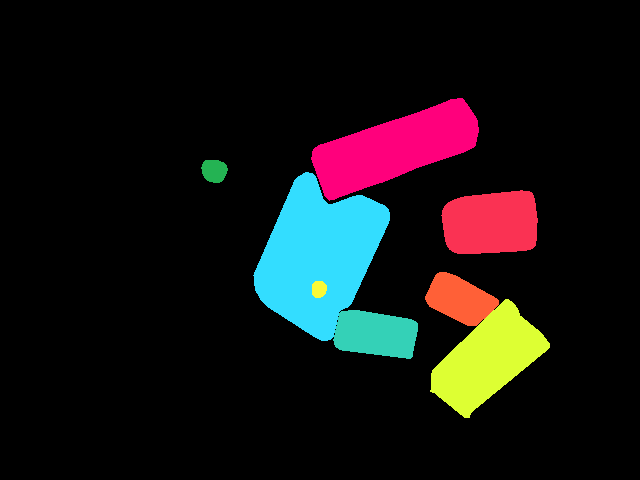}}
& \subfloat{\includegraphics[width = 1 in,height= 1 in]{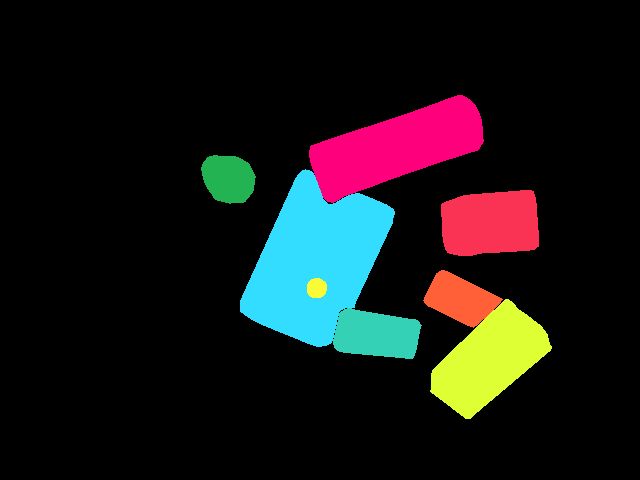}}\\
\hline

\rotatebox{90}{ESD-1 :Exp-3}
\rotatebox{90}{Rotational Motion }
&\subfloat{\includegraphics[width = 1 in,height= 1 in]{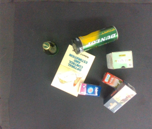}}  
& \subfloat{\includegraphics[width = 1 in,height= 1 in]{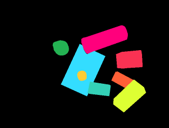}}
& \subfloat{\includegraphics[width = 1 in,height= 1 in]{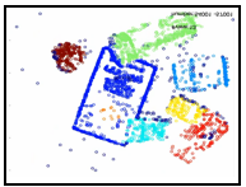}}
& \subfloat{\includegraphics[width = 1 in,height= 1 in]{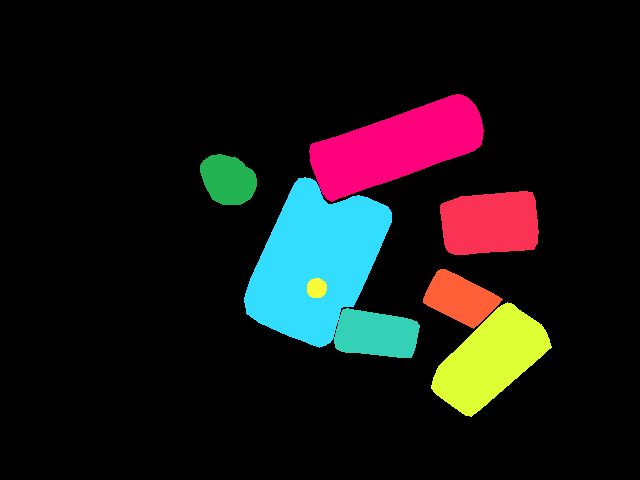}}
& \subfloat{\includegraphics[width = 1 in,height= 1 in]{Images/Light_Ours.png}}\\
\hline

\rotatebox{90}{ESD-1 : Exp-4}
\rotatebox{90}{0.3 m/s Speed}
&\subfloat{\includegraphics[width = 1 in,height= 1 in]{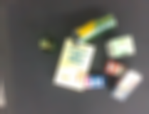}}  
& \subfloat{\includegraphics[width = 1 in,height= 1 in]{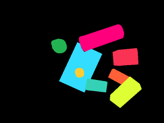}}
& \subfloat{\includegraphics[width = 1 in,height= 1 in]{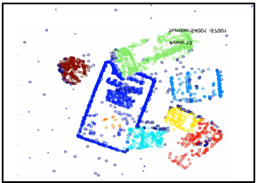}}
& \subfloat{\includegraphics[width = 1 in,height= 1 in]{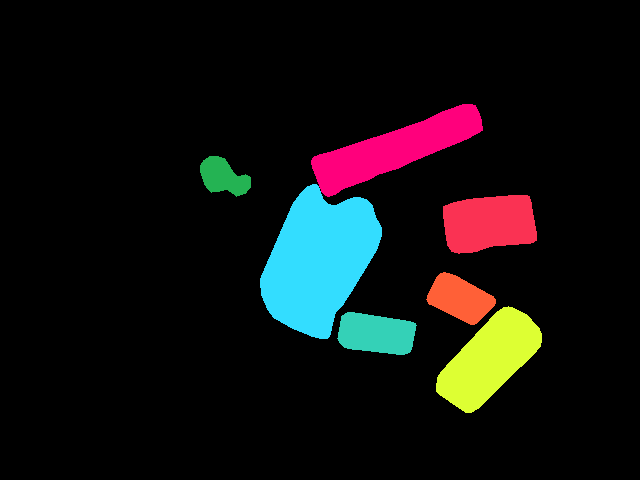}}
& \subfloat{\includegraphics[width = 1 in,height= 1 in]{Images/Light_Ours.png}}\\
\hline

\rotatebox{90}{ESD-1 : Exp-5 }
\rotatebox{90}{62cm Height }
&\subfloat{\includegraphics[width = 1 in,height= 1 in]{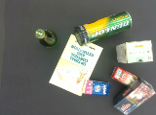}}  
& \subfloat{\includegraphics[width = 1 in,height= 1 in]{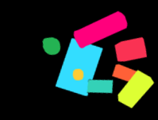}}
& \subfloat{\includegraphics[width = 1 in,height= 1 in]{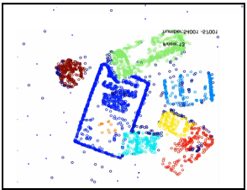}}
& \subfloat{\includegraphics[width = 1 in,height= 1 in]{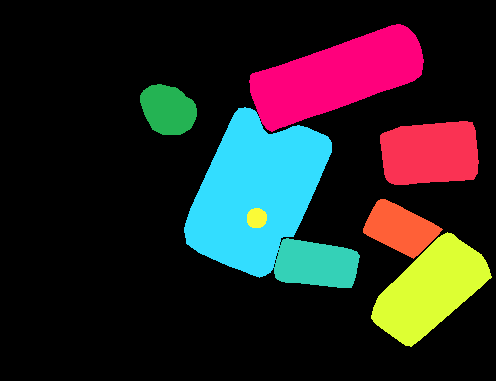}}
& \subfloat{\includegraphics[width = 1 in,height= 1 in]{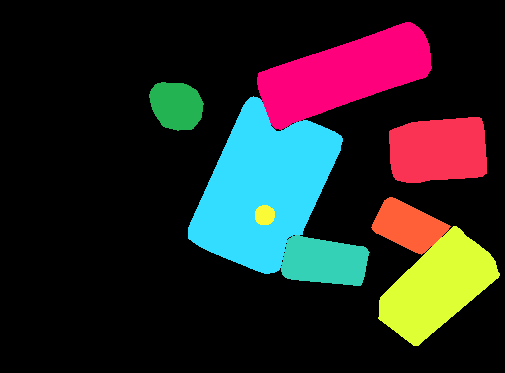}}\\
\hline

\rotatebox{90}{ESD-2 : Exp-3}
\rotatebox{90}{Rotational Motion}
&\subfloat{\includegraphics[width = 1 in,height= 1 in]{Images/Unknown_2_Obj_RGB.png}}  
& \subfloat{\includegraphics[width = 1 in,height= 1 in]{Images/Unknown_2_Obj_RGB_GT.png}}
& \subfloat{\includegraphics[width = 1 in,height= 1 in]{Images/Unknown_2_Obj_Event_GT.png}}
& \subfloat{\includegraphics[width = 1 in,height= 1 in]{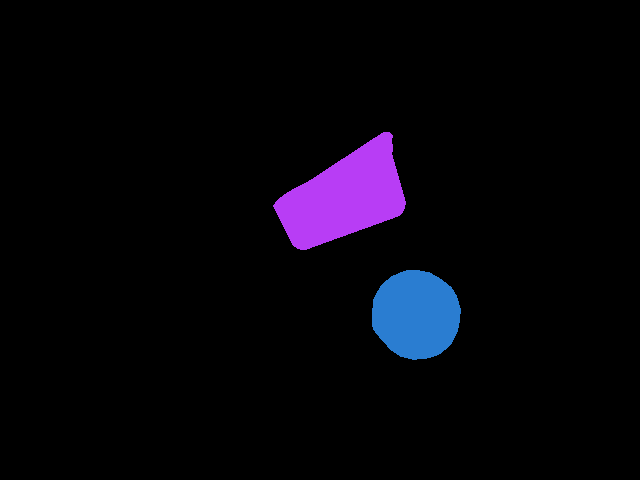}}
& \subfloat{\includegraphics[width = 1 in,height= 1 in]{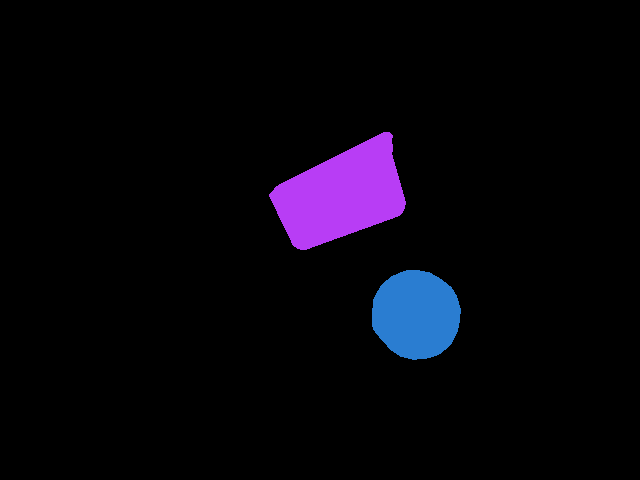}}\\
\hline

\rotatebox{90}{ESD-2 : Exp-1}
\rotatebox{90}{ Occlusion }
&\subfloat{\includegraphics[width = 1 in,height= 1 in]{Images/Unknown_5_Obj_RGB.png}}  
& \subfloat{\includegraphics[width = 1 in,height= 1 in]{Images/Unknown_5_Obj_RGB_GT.png}}
& \subfloat{\includegraphics[width = 1 in,height= 1 in]{Images/Unknown_5_Obj_Event_GT.png}}
& \subfloat{\includegraphics[width = 1 in,height= 1 in]{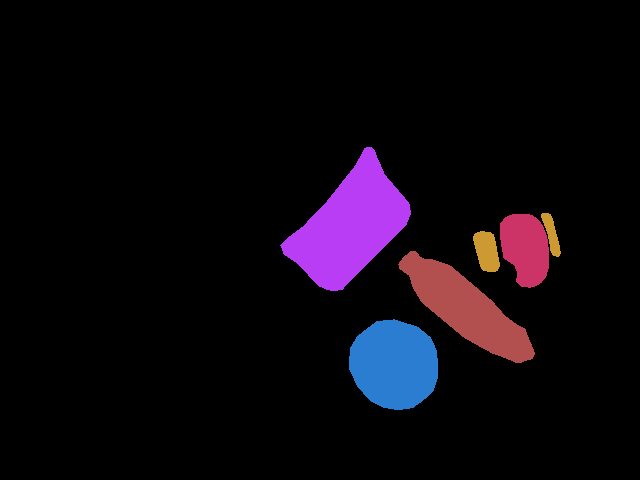}}
& \subfloat{\includegraphics[width = 1 in,height= 1 in]{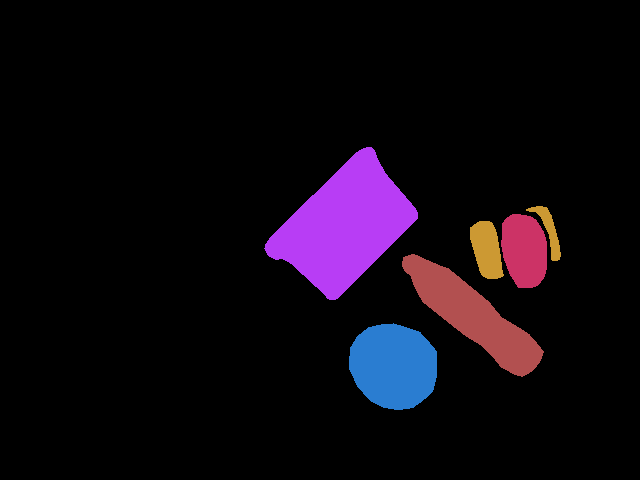}}\\
\hline

\end{tabular}
\end{adjustbox}

\caption{Qualitative Results - The qualitative results presented  compares the performance of four different methods, mainly CMX and (ours) for instance segmentation. The predictions were made on an ESD - 1  i.e. known objects and ESD -2 dataset ie. Unknown objects depicting the experiments conducted in quantitative evaluation.}
\label{fig:Qual Results}
\end{figure*}

In Fig. \ref{fig:Qual Results}, we present a variety of illustrative qualitative results, aiming to enhance the comprehension of our segmentation findings. Using ResNet-101 as the backbone, we have visually represented RGB-event segmentation results.

In the case of the ESD-1 dataset, which contains known objects, Bimodal SegNet effectively utilizes geometric information to correctly discern the contours of objects, displaying superior accuracy in comparison to the CMX model which exhibits deficiencies in contour precision. In the context of high-speed conditions, our methodology distinctly demarcates the boundaries between the box and its background, demonstrating an improvement in segmentation clarity. During the examination of the ESD-2 dataset, under challenging scenarios of occlusion conditions, our model was able to parse the mouse and box areas by maintaining satisfactory performance against the competitors. Additionally, the segmentation of the book's boundaries was achieved completely and smoothly, with distinctly delineated borders. The implementation of the AFPA module was found to yield positive effects, particularly observable in bottle detection tasks. In scenarios involving rotational motion, our model exhibits commendable generalization capabilities, leading to an enhancement in the segmentation of a moving camera. The addition of spatial information from an event-based vision modality contributed significantly to the accurate segmentation of objects. 

Summarizing the qualitative investigation, reinforces the applicability of our general approach to a wide range of dual-modal sensing combinations, facilitating robust instance scene understanding.

\section{{CONCLUSIONS}}
\label{section: Conclusion}

In this paper, Bimodal SegNet, a novel architecture for instance segmentation in robotic grasping using a dynamic vision sensor is presented. The architecture fuses the event and RGB frames at various resolutions of the event encoder and RGB encoder using a cross attention mechanism, which further fuses the feature maps in the decoder. The feature maps from both encoders are weighted and passed into the APFA block where atrous convolutions are employed.

Our proposed model achieves state-of-the-art performance on the ESD-1 and ESD-2 datasets in terms of mIoU and pixel accuracy, as demonstrated in multiple scenarios with variations in object types, trajectory, camera speed, apparent size, and lighting conditions. The fusion approach we propose achieves high robustness against various challenges, such as occlusions, low lighting, small objects, high speed, and linear motion. Bimodal SegNet, leveraging the complementarity of events and frames, outperforms unimodal encoder-decoder approaches. In comparison to the state-of-the-art transformer-based CMX, the proposed Bimodal SegNet exhibits higher accuracy, with lower complexity and computational cost, and therefore the superiority of our proposed architecture that fuses RGB and event frames at multiple resolutions, with skip connections and atrous convolutions is demonstrated.

In the future, we aim to enhance the capabilities of the event camera by integrating an asynchronous event-based approach. This would effectively leverage the high temporal resolution of the event camera for segmenting unknown objects in robotic grasping. Moreover, this advancement could be generalized to a variety of tasks in industrial applications, such as real-time identification and manipulation on assembly lines, robotics-based sorting, and machining tasks such as drilling and sinking to establish more robust and adaptive machine vision systems.

\section*{{Acknowledgments}}{This work was supported by Kingston University, the Advanced Research and Innovation Center (ARIC) and Khalifa University of Science and Technology, Abu Dhabi, UAE.}

\bibliographystyle{IEEEtran}

\bibliography{references}


\end{document}